\documentclass[sigconf]{acmart}
\usepackage{graphicx}
\usepackage{amsthm}
\usepackage{multirow}
\usepackage{subfigure}
\usepackage{enumitem}
\usepackage{xcolor}
\usepackage{xspace}

\newcommand{\eg}{\emph{e.g.},\xspace}

\newcommand{\ie}{\emph{i.e.},\xspace}

\newcommand{\eat}[1]{}

\newtheorem{problem}{Problem}
\newtheorem{myDef}{Definition}[]
\newtheorem{mythe}{theorem}[]
\newtheorem{mylem}{Lemma}[]

\AtBeginDocument{%
  \providecommand\BibTeX{{%
    \normalfont B\kern-0.5em{\scshape i\kern-0.25em b}\kern-0.8em\TeX}}}

\setcopyright{acmcopyright}
\copyrightyear{2023}
\acmYear{2023}
\acmDOI{XXXXXXX.XXXXXXX}
 
\acmConference[Conference acronym 'XX]{Make sure to enter the correct
  conference title from your rights confirmation emai}{June 03--05,
  2018}{Woodstock, NY}

\acmPrice{15.00}
\acmISBN{978-1-4503-XXXX-X/18/06}

\usepackage{xcolor} 

\begin{document}
\title{HSTFL: A Heterogeneous Federated Learning Framework for Misaligned Spatiotemporal Forecasting}
\author{Shuowei Cai}
\affiliation{%
  \institution{Artificial Intelligence Thrust, The Hong Kong University of Science and Technology (Guangzhou)}
    \country{}
}
\email{scaiak@connect.hkust-gz.edu.cn}
\author{Hao Liu}

\affiliation{%
  \institution{Artificial Intelligence Thrust, The Hong Kong University of Science and Technology (Guangzhou)\\
  Department of Computer Science and Engineering, The Hong Kong University of Science and Technology}
  \country{}
}
\email{liuh@ust.hk}

\begin{abstract}
Spatiotemporal forecasting has emerged as an indispensable building block of diverse smart city applications, such as intelligent transportation and smart energy management. Recent advancements have uncovered that the performance of spatiotemporal forecasting can be significantly improved by integrating knowledge in geo-distributed time series data from different domains, \eg enhancing real-estate appraisal with human mobility data; joint taxi and bike demand predictions. While effective, existing approaches assume a centralized data collection and exploitation environment, overlooking the privacy and commercial interest concerns associated with data owned by different parties. In this paper, we investigate multi-party collaborative spatiotemporal forecasting without direct access to multi-source private data. However, this task is challenging due to 1) cross-domain feature heterogeneity and 2) cross-client geographical heterogeneity, where standard horizontal or vertical federated learning is inapplicable. To this end, we propose a Heterogeneous SpatioTemporal Federated Learning (HSTFL) framework to enable multiple clients to collaboratively harness geo-distributed time series data from different domains while preserving privacy. 
Specifically, we first devise vertical federated spatiotemporal representation learning to locally preserve spatiotemporal dependencies among individual participants and generate effective representations for heterogeneous data. Then we propose a cross-client virtual node alignment block to incorporate cross-client spatiotemporal dependencies via a multi-level knowledge fusion scheme. Extensive privacy analysis and experimental evaluations demonstrate that HSTFL not only effectively resists inference attacks but also provides a significant improvement against various baselines.
\end{abstract}

\maketitle
\section{Introduction}
The proliferation of Internet of Things (IoT) devices has generated massive geo-tagged time series data in urban spaces, such as travel demands~\cite{st-resnet} and user check-ins~\cite{checkins}. 
Spatiotemporal forecasting aims at predicting the future states of geo-distributed time series based on their historical data, which has become a cornerstone for various smart city applications, \eg transportation optimization, urban planning, and energy management. 
Therefore, extensive efforts have been made for different spatiotemporal forecasting tasks ~\cite{spatiotemporal,urbancomputing}, where the common routine is capturing complicated spatial and temporal correlations via deep neural networks, \eg Convolution Neural Networks (CNNs), Graph Neural Networks (GNNs), and Recurrent Neural Networks~(RNNs)~\cite{st-resnet,stgcn,gwnet,autocts}.

\begin{figure}
  \centering
  \includegraphics[width=0.46\textwidth]{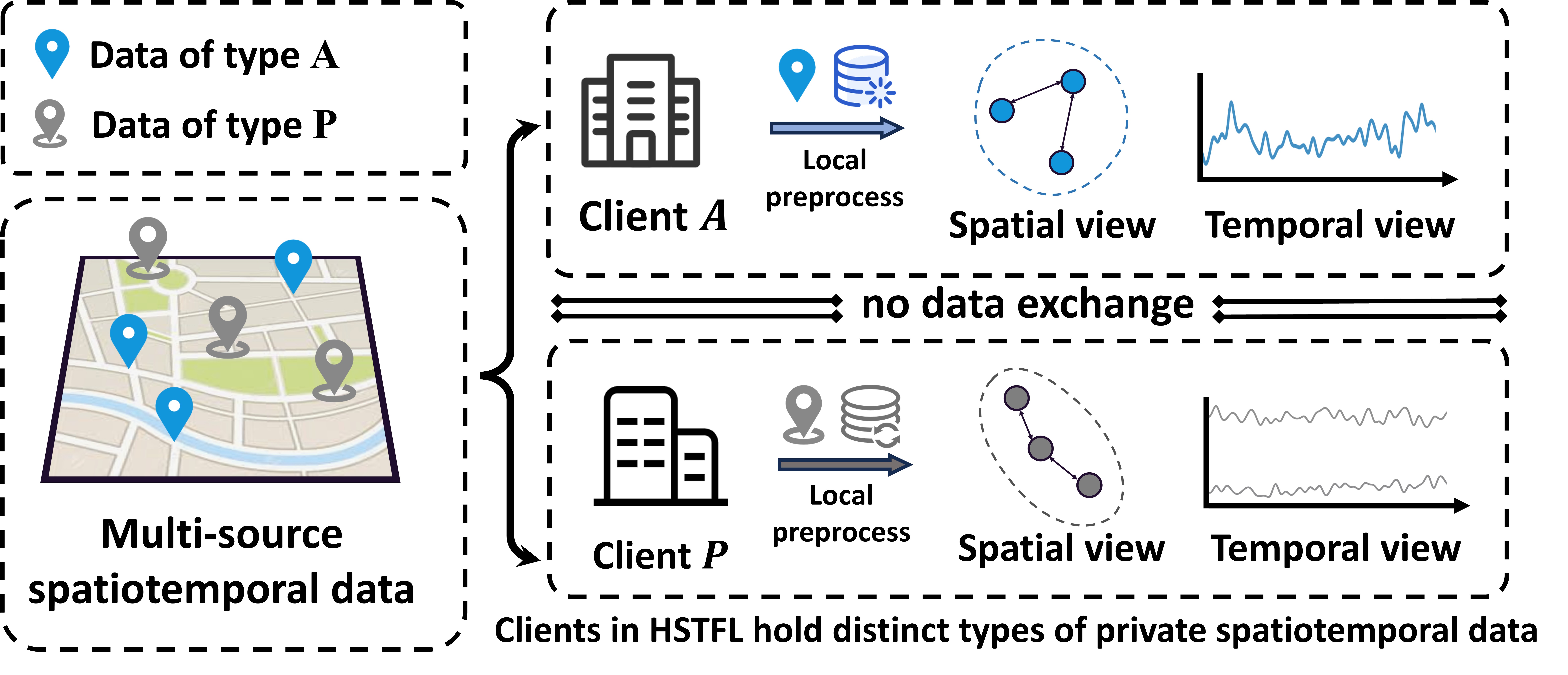}
  \caption{
  Illustrative example of multi-source private spatiotemporal data hold by multiple clients. 
  Each client holds a distinct type of private geo-distributed time series (\ie spatiotemporal data). These geo-distributed time series data in each clients are also misaligned in geographical location.}
  \label{fig:hstdata}
  \vspace{-0.6cm}
\end{figure}

Recent advancements reveal the effectiveness of spatiotemporal forecasting can be substantially improved by sharing information from geo-distributed time series data across various domains. To name a few, 
CoST-Net~\cite{costnet} constructs a co-prediction framework to mutually reinforce taxi and bike demand prediction tasks by exploiting macro and micro knowledge among taxi and bike demand time series, while MasterGNN~\cite{mastergnnc} devises a multi-adversarial network to harness the inner-connection of air quality and weather condition data.
As another example, MugRep~\cite{mugrep} enhances the real estate appraisal accuracy by introducing time-varying human mobility information from massive user trip and check-in data.
However, existing approaches typically assume the multi-source spatiotemporal data can be collected and processed in a centralized way, which is not always practical in the real world.
On the one hand, the multi-source data held by different parties may contain immense commercial value. For example, the taxi demand data held by a ride-sharing platform can be leveraged to improve its competitors’ dispatching strategy~\cite{fedltd}.
On the other hand, the spatiotemporal data usually involve confidential information, \eg private user locations in taxi demand and check-in data~\cite{mugrep}.
As shown in Figure~\ref{fig:hstdata}, data holders are unable to disclose their private spatiotemporal data because of commercial interest or regulation constraints.
To bridge this gap, this work investigates multi-party collaborative spatiotemporal forecasting without directly accessing multi-source urban data.

Federated learning (FL) has emerged as a promising solution to leverage distributed data while preserving privacy, enabling multiple parties to collaboratively construct a model by exchanging intermediate results instead of exposing raw data~\cite{fedavg,advance_fl}. However, conventional FL approaches encounter substantial challenges when applied to spatiotemporal forecasting with multi-source urban data, primarily due to two factors.

(1)~\textit{Cross-domain feature heterogeneity}: Geo-distributed time series data collected by different parties often exhibit considerable heterogeneity, such as varying dimensions and feature types. 
For example, the real estate appraisal data may consist of historical transaction details such as the transaction date, price, location, room number, and transaction ownership, while the human mobility data mainly include the visiting frequency, volume, and travel modes in each time slots~\cite{mugrep,humanmobility}.  Existing spatiotemporal federated learning methods ~\cite{cnfgnn,stfl,pfedweather}, predominantly following the Horizontal Federated Learning (HFL) paradigm~\cite{fl_qiang}, assume homogeneous data types across different regions. While adept at extracting common knowledge from similar data distributions, such HFL-based methods are inherently unable to handle heterogeneous features in multi-source spatiotemporal data collected from different domains. Therefore, how to collaboratively learn a joint spatiotemporal forecasting model that can effectively handle such feature heterogeneity across different domains is the first challenge.

(2)~\textit{Cross-client geographical heterogeneity}: The time series data collected from different parties are not only heterogeneous in feature space but also in their geospatial distribution. While Vertical Federated Learning (VFL)~\cite{vflsurvey} has shown promise in handling data of different types and varying dimensions, existing VFL algorithms~\cite{fdml,vsplit,vfgnn} struggle with the complex correlations among time series data that are geographically misaligned. More in specific, VFL frameworks typically rely on entity alignment~\cite{recordlinkage} to link data features across multiple clients, such as matching records of the same user. However, such an alignment process is less feasible for geographically misaligned spatiotemporal data. Although a few studies, such as fuzzy linkage~\cite{fedsim}, attempt to define correlations between misaligned data in VFL, they overly rely on heuristic rules and may lose critical cross-client spatiotemporal dependencies. Moreover, such VFL methods often expose data representations during cross-client information exchange, increasing the risk of privacy leakage. How to share cross-client spatiotemporal knowledge based on geographically misaligned time series with privacy preservation is another challenge.

To address the above challenges, we propose the \textbf{H}eterogeneous \textbf{S}patio\textbf{T}emporal \textbf{F}ederated \textbf{L}earning~(HSTFL) framework to facilitate effective and privacy-preserving multi-party collaborative spatiotemporal forecasting.
Specifically, we first propose vertical federated spatiotemporal representation learning that enables clients to generate spatiotemporal representations for data with feature heterogeneity. It consists of temporal representation learning and vertical federated spatial representation learning for capturing intra-client temporal and spatial correlations, respectively.
Moreover, we propose a cross-client virtual node alignment block to align geographically heterogeneous time series data and facilitate cross-client correlation propagation in a privacy-preserving manner.
In particular, a set of virtual nodes is generated and fused through a privacy-preserving multi-level alignment scheme to incorporate cross-client spatiotemporal dependencies. 
To ensure the privacy of clients' data, we adapt a privacy-preserving framework construction with no-model-sharing design, incorporate differential privacy against attacks and reduce prior knowledge leakages in HSTFL.

In summary, this work makes the following contributions.
\begin{enumerate}
\item To the best of our knowledge, this work is the first attempt to investigate multi-party collaborative spatiotemporal forecasting without direct access to heterogeneous spatiotemporal data. We identify the feature heterogeneity and geographical heterogeneity challenges raised in federated spatiotemporal forecasting with multi-source data. 
\item We propose the Heterogeneous SpatioTemporal Federated Learning framework. By integrating vertical federated spatiotemporal representation learning with cross-client virtual node alignment, it effectively utilizes multi-source spatiotemporal data for accurate forecasting while protecting the privacy of each participant.
\item Extensive experiments on four real-world multi-source urban spatiotemporal datasets show that HSTFL can effectively enhance prediction effectiveness against local models and existing federated learning baselines. Moreover, we conduct a comprehensive theoretical and empirical privacy analysis of our proposed framework, proving its effectiveness in resisting various attack algorithms.
\end{enumerate}

\section{Related Work}
\subsection{Federated Learning on Spatiotemporal Data}
Various FL algoithms have been proposed for privacy-preserving collaborative learning on spatiotemporal data~\cite{cnfgnn,pfedweather,stfl,fedstn,Onlinest}. To name a few, CNFGNN is a cross-device FL algorithm that enables clients to model spatiotemporal correlation by passing representations of local time series to GNNs at server while training with FedAVG~\cite{cnfgnn}; MetePFL explores a foundation-model-based solution for privacy-preserving cross-silo weather forecasting in different regions with prompt federated learning~\cite{pfedweather}; STFL transforms time series data into graph-structured data, and trains GNNs for cross-silo FL~\cite{stfl}. However, existing spatiotemporal FL algorithms assume that clients hold data of the same type. They are inapplicable for the scenarios that clients hold completely different types of data.

\subsection{Federated Learning on Heterogeneous Data}
Federated learning involves clients with diverse characteristics, resulting in potential variations in data held by each client. To address the challenge of data heterogeneity, various HFL algorithms such as clustered federated learning~\cite{clusteredFL} and personalized FL ~\cite{personalizedFL} are proposed. However, these methods still aim to extract common knowledge from similar data distribution in multiple clients and necessitate an aggregation of clients' local model, hence inherently unable to handle the scenario in which participants hold distinct types of data with different feature dimensions. 

To accommodate the data with strong heterogeneity, VFL~\cite{vflsurvey} enables clients to train data of different types with privacy-preserving entity alignment~\cite{recordlinkage}. Privacy-preserving entity alignment match the data samples with their features from different clients to the same entities in VFL, (\eg matching user records in multiple clients to the user with the user id), thus providing additional information to the entities to enhance predictions with VFL models ~\cite{VLR,sbt, fedsvd,vfgnn}.
However, exact alignment may not always be applicable in VFL. To address this issue, FedSim~\cite{fedsim} proposed fuzzy linkage to match the cross-client data with common features and designed a similarity-based VFL with SplitNN~\cite{vsplit}.
However, it fails to comprehensively model the complex correlation between clients' data and introduces additional privacy leakage with the fuzzy linkage.

\subsection{Multi-source Data for Urban Spatiotemporal Forecasting}
Research has been conducted to integrate data from multiple domains to construct multi-source datasets for effective spatiotemporal forecasting in urban computing~\cite{urban-fusion}.
Additional data sources such as check-in records, POIs, weather, taxi and traffic flow are used to enhance predictions in various domains, including air quality, bike demand, parking, and real estate appraisal \cite{geoman, u-air,mastergnnj,mastergnn3, costnet, coevolve, mdtp, jointparking, mugrep}. 
Methodologically, these efforts employ a centralized preprocessing for heterogeneous spatiotemporal data (\eg converting geo-distributed data into city grids ~\cite{costnet,coevolve} and matching similar time series~\cite{mastergnnc}). Subsequently, task-specific spatiotemporal models with stacked spatiotemporal modules are trained using the processed multi-source spatiotemporal data. Nevertheless, existing works fail to address the privacy concern of such multi-source spatiotemporal data as they are naturally collected and preprocessed by different companies and institutions. The fail to provide a solution for collaborative spatiotemporal forecasting without direct access to multi-source private spatiotemporal data.

\section{Preliminary}
\subsection{Problem Formulation}

In this work, each client (\ie participant)  holds a unique type of time series data, with the data types varying among different clients. We categorize the clients into two types of roles.
\textit{Active party} $A$ is the client who is interested in constructing a forecasting model. It holds spatiotemporal data within a specific domain along with the target labels (\eg bike demand and property prices).
\textit{Passive party} $P$ is the client that holds spatiotemporal data in a domain different from that of the active party (\eg taxi demand), which potentially enhances the spatiotemporal forecasting performance for $A$. 
Without loss of generality, in this work, we assume each collaborative learning task involves only one active party.

Given the active party $A$, let $X^{A} = \{ x^{a_i} \}_{i=1}^{N^A}$ represent the time series data of the active party, where $x^{a_i}$ is a feature vector associated with a location $a_i$. We use $X^{A,T_A}$ to denote the $T_A$-time step series of $N^A$ locations held by party $A$.
Similarly, given multiple passive parties $\{P_1, \ldots, P_J\}$, for each passive party $P_j$, their geo-distributed time series data are denoted as $X^{P_j,T_j} = \{ x^{p_j,T_j}_i \}_{i=1}^{N^{P_j}}$. 
It's important to note that these time series data may vary in data types, feature dimensions, sampling rate and time steps~(\ie $T_A \neq T_j$), leading to huge heterogeneity across clients. Let $\mathbf{X}=\{X^{A,T}\}\cup \{X^{P,T}\}^T_{t \in T}$ denote heterogeneous spatiotemporal data features of all parties in previous time steps, we define the problem as follows.

\begin{problem}
\textbf{Privacy-preserving Heterogeneous Spatiotemporal Forecasting}. Given the active party $A$ and the passive parties $P=\{P_1,...,P_J\}$, along with their heterogeneous time series $\mathbf{X}$. Our goal at a given time step $t$ is to predict the future states of all locations in $A$ over the next $\tau$ time steps
\begin{equation}
(\hat{\mathbf{Y}}^{A,t+1},\hat{\mathbf{Y}}^{A,t+2},...,\hat{\mathbf{Y}}^{A,t+\tau}) \leftarrow \mathcal{M}(\mathbf{X}),
\end{equation}
where $\hat{\mathbf{Y}}^{A,t+1}$  is the predicted outcomes for all time series in the active party at time step $t+1$. $\mathcal{M}$ is the mapping function we aim to learn collaboratively with privacy preservation.
\end{problem}


\subsection{Threat Model}
In this paper, we consider the adversary to be semi-honest (honest-but-curious) like~\cite{secureml,VLR,vfgnn}. Under this model, while the clients adhere strictly to the prescribed federated learning protocol for training, they may attempt to extract as much information as possible from the shared intermediate results.
Specifically, these semi-honest clients may employ inference attacks like~\cite{FIattack} on the intermediate results with prior knowledge to reconstruct sensitive time series data of other clients as much as possible.

\section{Heterogeneous Spatiotemporal Federated Learning}

\subsection{Overview} 
\begin{figure}
  \setlength{\abovecaptionskip}{0.2cm}
  \centering
  \includegraphics[width=0.43\textwidth]{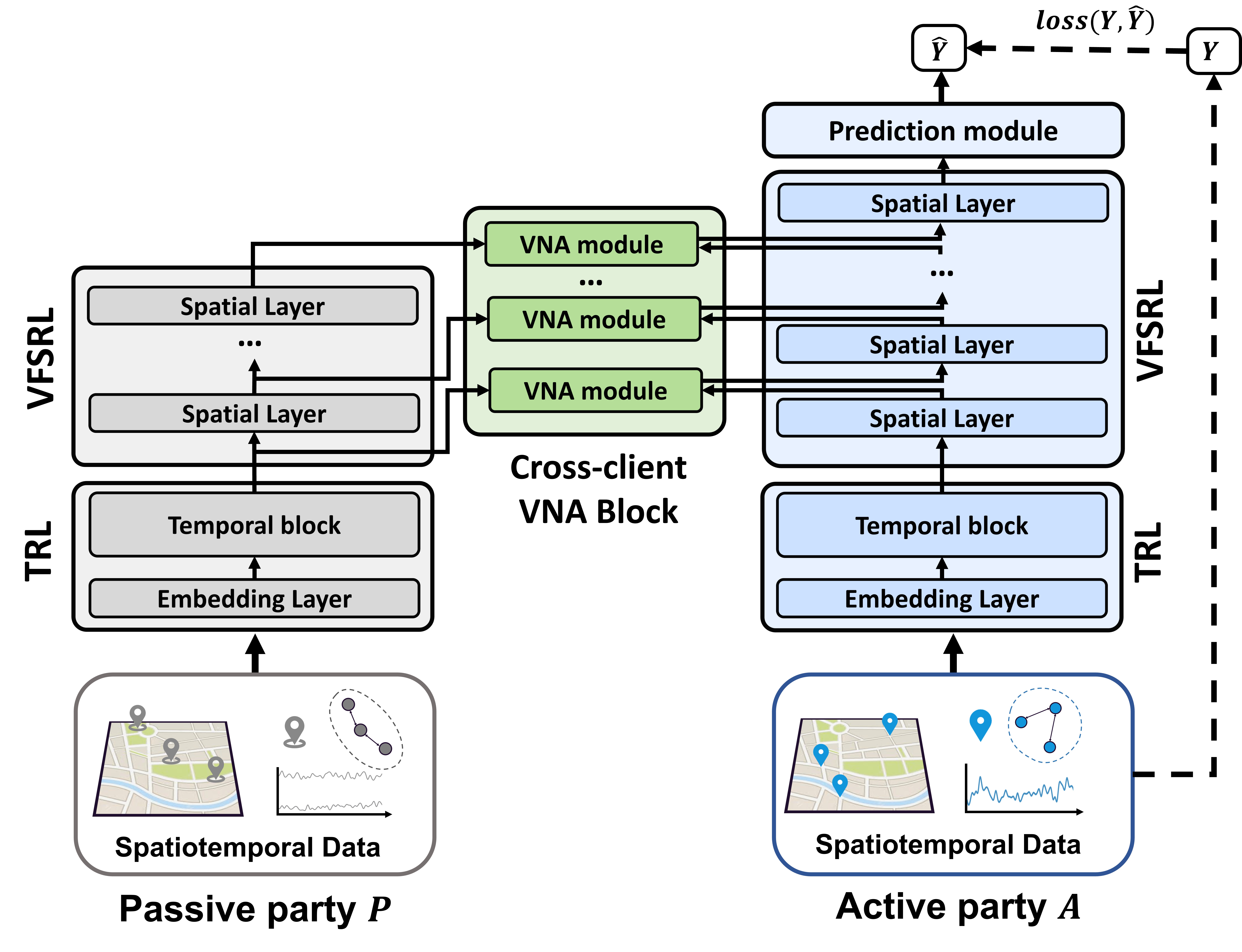}
  \caption{Overview of the HSTFL framework.}
  \label{fig:hstfl}
  \vspace{-0.5cm}
\end{figure}

Figure~\ref{fig:hstfl} shows an overview of HSTFL, which includes two major privacy-preservation tasks:
(1) capturing temporal dependency in each heterogeneous time series and local spatial dependency of geo-distributed observations in each client;
(2) modeling cross-client spatiotemporal dependency among geographically misaligned time series.
In the first task, we employ Vertical Federated SpatioTemporal Representation Learning (VFSTRL), with Temporal Representation Learning (TRL) to project heterogeneous and length-varying time series features into low-dimensional embeddings with the preservation of temporal dependency, as well as Vertical Federated Spatial Representation Learning (VFSRL) to extract spatial dependencies of geo-distributed time series and generate effective representations in each client.
In the second task, to further incorporate cross-client dependencies with privacy-preserving, we propose a cross-client Virtual Node Alignment (VNA) block to enable knowledge propagation between spatially correlated time series held by different clients with virtual node generation, privacy-preserving knowledge fusion for cross-client multi-level alignment. 
Overall, the proposed framework follows a temporal-then-spatial architecture for spatiotemporal forecasting~\cite{tng,TGAT,lightcts}. It is noteworthy that HSTFL adapts a no-model-sharing design, where no model parameter is shared among clients and all parties use their local private model for prediction. Since all passive parties has the same procedure, we only demonstrate one passive party in the following.

\subsection{Vertical Federated Spatiotemporal Representation Learning}
This section introduces vertical federated spatiotemporal representation learning, which generate multi-level spatiotemporal representations of clients' data with TRL and VFSRL.
\subsubsection{Temporal Representation Learning}
Clients in HSTFL hold complete local time series. Therefore, we deploy TRL at first, enabling clients to conduct temporal correlation modeling locally and project length-varying time series features to unified representations with their local temporal model. The active party and the passive parties follow the same process in TRL, and the time series within each client share the same local temporal model.

Specifically, the local temporal model in TRL consists of an embedding layer and a temporal block with stacked temporal layers. For time series data $x^{c_i} \in \mathbb{R}^{T\times F^C}$ in a client $C$  ($C$ could be the active party or a passive party) with feature dimension $F^C$, the embedding layer maps it to a latent representation $h_{\mathscr{t},0}^{c_i} \in \mathbb{R}^{T\times H}$ with hidden units $H^C_{t} > F^C$. 
Then, a temporal block with $M_t$ stacked temporal layers (\eg GRU, LSTM, Informer\cite{informer}) is constructed to capture the temporal correlations and generate the temporal representation $h_{T}^{c_i} \in \mathbb{R}^{H}$,
\begin{equation}
\begin{aligned}
h_{\mathscr{t},0}^{c_i}&=EmbeddingLayer(x^{c_i}),\\
h_{\mathscr{t},m}^{c_i}&=TemporalLayer_m(h_{\mathscr{t},m-1}^{c_i}) , m=1,...,M_t.
\end{aligned}
\end{equation}
The modules for the temporal layers can be selected based on local data characteristics. We use a linear layer for the embedding layer and stacked GRU for the temporal layers by default. The last time step in $h_{\mathscr{t},M_t}^{c_i}$ serve as the temporal representation $h_{T}^{c_i}$.

\subsubsection{Vertical Federated Spatial Representation Learning}
Vertical federated spatial representation learning models the intra-client spatial correlations based on the temporal representation and generates spatiotemporal representations for cross-client spatial correlation modeling with local spatial model. The local spatial model in VFSRL of a client $C$ is made up of $M_s$ stacked spatial layers, \ie
\begin{equation}
\begin{aligned}
o_{m}^C=SpatialLayer_m(h_{\mathscr{s},m-1}^C), m=1,...,M_s.
\end{aligned}
\end{equation}

$h_{\mathscr{s},m-1}^C$ and $o_{m}^C$ represent the input and output of the $m-th$ spatial layer $SpatialLayer_m$ for client $C$, where $h_{\mathscr{s},0}^C$ is the temporal representation of $C$ (spatial information for spatial layers are omitted for simplicity). In practice, the spatial layer can take the form of CNN, GNN and their variants according to the data characteristics. We use the spatial layer of Graph WaveNet (GWN)~\cite{gwnet} by default.
 
Compared to existing vertical federated deep learning algori-thms ~\cite{fdml,vsplit,vfgnn} that directly apply the output of the topmost layer in client's private model as representation, HSTFL utilizes the outputs in each spatial layer with the temporal representation to construct a set of multi-level representations as the spatiotemporal representation, denoted as $h_{ST}^C= \{ h_{T}^C\}  \cup  \{ o_{m}^C |m=1,...,M_s \}$. This multi-level representation enables HSTFL to capture comprehensive cross-client spatial correlations in the cross-client virtual node alignment block, which will be detailed in Section \ref{section:multi-level}.

The way to prepare input for the next spatial layers differs based on the client's role. For the passive party $P$, the output $o_{m}^P$ will be fed into the next spatial layer as $h_{\mathscr{s},m}^P$ directly. The passive parties will first generate the spatiotemporal representation and then pass it to the cross-client VNA block. As for the active party, it receives information from the cross-client VNA block at first, and then updates its local representations $o_{m}^A$ to obtain $h_{\mathscr{s},m}^A$ as the input for the next spatial layer. Finally, the active party applies an MLP as prediction module with $o_{M_s}^C$ to generate $(\hat{\mathbf{Y}}^{A,t+1},\hat{\mathbf{Y}}^{A,t+2},...,\hat{\mathbf{Y}}^{A,t+\tau})$ as the prediction.

\subsection{Cross-Client Virtual Node Alignment}
\label{section:vna}
\begin{figure}
  \centering
  \includegraphics[width=0.44\textwidth]{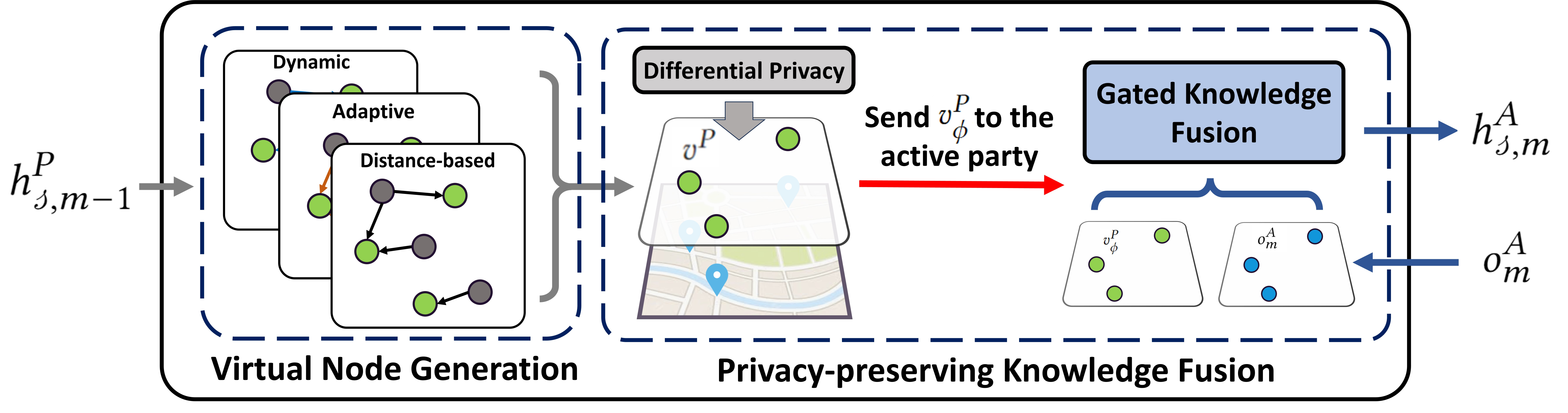}
  \caption{The virtual node alignment module.}
  \label{fig:vna}
\end{figure}

The cross-client virtual node alignment block conducts cross-client correlation modeling between geo-distributed time series with the collaboration of the active party and the passive parties. 
As shown in Figure~\ref{fig:vna}, a VNA module consists of virtual node generation and privacy-preserving knowledge fusion.

\subsubsection{Virtual Node Generation}
The virtual nodes are a set of virtual entities that are generated by the passive party for aligning misaligned time series.
The virtual nodes correspond one-to-one with the time series in the active party, having the same quantity and geographical coordinates as them for alignment.
The passive party $P$ generates virtual nodes from its spatiotemporal representation. 
Given a single-level representation $z^P \in \mathbb{R}^{N^P\times H^P}$ from the set of multi-level spatiotemporal representation $h_{ST}^P$, the virtual nodes $v^P \in \mathbb{R}^{N^A\times H^A}$ are generated with a hybrid aggregation method to model the complex cross-client spatiotemporal correlations.

\textbf{Distance-based aggregation}.
We first leverage the distance information to aggregate spatially proximate information using K-nearest neighbors. Given the distance matrix $\Delta\in \mathbb{R}^{C^P\times C^A}$ which contains the geographic distance between the party's time series and the virtual nodes, the KNN function sets the elements to 1 if its corresponding distance is the smallest $k$ values from the virtual node to a time series in the passive party, and 0 otherwise. Then, a linear layer $W_{dis}$ is applied to aggregates these information, 
\begin{equation}\label{eq:dis}
v_{dis}= W_{dis} (KNN(\Delta,k) z^P).
\end{equation}
Compared with fuzzy linkage~\cite{fedsim}, distance-based aggregation does not require the exposure of the locations (\ie common features in fuzzy linkage) of the time series, as both the distance matrix and the KNN matrix can be computed in advance with privacy-preserving protocols such as secret sharing~\cite{secureml}. Moreover, it only shares a projected virtual node rather than Top-$k$ nearest representations, therefore further reducing potential privacy leakage.
\textbf{Adaptive aggregation}: 
Beyond the spatial proximity, we further introduce the adaptive aggregation scheme to incorporate more comprehensive implicit correlations, \eg regional synchronization effects~\cite{MGchai,MGgeng}.
Specifically, we adapt two learnable weights $W_{A1} \in \mathbb{R}^{C^P\times d}$ and $W_{A2} \in \mathbb{R}^{C^A\times d}$ with Relu and Softmax to construct a self-adaptive adjacency matrix~\cite{gwnet}. 
Then, a linear layer $W_{adp}$ is adapt to aggregate information with the matrix,
\begin{equation}
v_{adp}=W_{adp}(Softmax(Relu(W_{A1}W_{A2}^T))z^P).
\end{equation}
Adaptive aggregation can effectively address situations that significant disparities exist in the geo-distribution of clients' time series, where spatial proximity modeling is inadequate. Moreover, it can uncover long-range and implicit static correlations with a global view, to aggregate more comprehensive spatial information.

\textbf{Dynamic aggregation}: 
The correlations between time series can change significantly over time \cite{gman,lightcts}, which requires a dynamic correlation modeling to capture this correlation. We proposed a multi-attention mechanism to generate a dynamic attention matrix to model dynamic cross-client spatial correlations. 

First, we define trainable positional embeddings $X_{z^P} \in \mathbb{R}^{C^P\times H}$ and $X_{v^P} \in \mathbb{R}^{C^A\times H}$ for the time series and virtual nodes, respectively. Then, we combine the representation with the positional embedding to generate the input for calculating the attention matrix with $X_{in}=cat(\{z^P+X_{z^P},X_{v^P}\})$, $X_{out}=X_{v^P}$. By such a design, the attention matrix can be constructed based on the dynamic information of the time series and the static information of both the virtual nodes and the time series. Then, we derive the attention score,
\begin{equation}
Q_i=X_{out}W^Q_i,K_i=X_{in}W^K_i,V_i=X_{in}W^V_i, 
\end{equation}
\begin{equation}
A_{att}^i=softmax(\frac{Q_i \cdot K_i^T}{\sqrt{(C^P+C^A)/n_{head}}}),
\end{equation}
where $n_{head}$ is the number of attention head, $W^Q_i \in \mathbb{R}^{C^A\times\frac{(C^P+C^A)}{n_{head}}}$, $W^K_i,W^V_i \in \mathbb{R}^{(C^P+C^A)\times\frac{(C^P+C^A)}{n_{head}}}$ are three learnable matrices for attention head $i$.
We aggregate the information in a dynamic manner with attention matrices $cat(\{A_{att}^i \cdot V_i\}^{n_{head}}_{i=1})$ and two linear layers $W_2$ and $W_1$,
\begin{equation}
\begin{aligned}
v_{dyn}=W_2(ReLU(W_1 (cat(\{A_{att}^i \cdot V_i\}^{n_{head}}_{i=1})))).
\end{aligned}
\end{equation}

The above aggregation schemes model the static correlations and dynamic correlations between heterogeneous time series data and derive comprehensive information for prediction. Finally, we fuse the immediate representations to generate the virtual node,
\begin{equation}
\begin{aligned}
v^P=ReLU(v_{dis}+v_{adp}+v_{dyn}).
\label{section:relu}
\end{aligned}
\end{equation}

\subsubsection{Privacy-preserving Knowledge Fusion}
Privacy-preserving kn-owledge fusion enables the active party to obtain information from VNA modules in a privacy-preserving manner in VFSRL. Given the output $o_m^A$ of a spatial layer $m$ at the active party $A$ and the corresponding virtual node $v^P$ from the passive party $P$, privacy-preserving knowledge fusion outputs $h_{\mathscr{s},m}^A$ as the input of the next spatial layer. As shown in Figure~\ref{fig:vna}, privacy-preserving knowledge fusion consists of three steps: the passive party employs differential privacy to protect the virtual nodes, send virtual nodes to the active party and the active party conducts gated knowledge fusion.

\textbf{Differential privacy on virtual nodes.}
The virtual nodes generated by the passive parties need to be sent to the active party for knowledge fusion. To further reduce the information leakage in the virtual nodes, we incorporate differential privacy~\cite{dp} (DP) to safeguard data privacy.
\begin{myDef}[The Gaussian Mechanism]
    \textit{Given a function $f:D\to \mathbb{R}^d $ over a dataset $D$, the Gaussian mechanism is defined as }
    $$M_G(x,f(\cdot),\epsilon)=f(x)+(r_1,...,r_k).$$
    Where $r_i$ are $i.i.d$ random variables drawn from $\mathcal{N}(0,\sigma^2\Delta_2f^2) $  and $ \sigma=\frac{\sqrt{2ln(1.25/\delta)}}{\epsilon}$.
\label{def:gaussian}
\end{myDef}
\begin{mythe}
    \textit{ The Gaussian mechanism defined in Definition \ref{def:gaussian} preserves $(\epsilon ,\delta)$-DP for each publication step~\cite{dp}}.
\end{mythe}
Consider the passive party's private model as a function $f(\cdot)$. The virtual nodes are $f(x)$ with private data $x$ as input. We calculate the $l_2$-sensitivity with norm clipping to apply the Gaussian mechanism on the virtual node $v^P$ to obtain the protected virtual nodes $v_{\phi}^P$. The protected virtual nodes will be sent to the active party. 

\textbf{Gated knowledge fusion.} After the active party received the protected virtual node $v_{\phi}^P$, a gated fusion is conducted to selectively integrate information with its own representations,
\begin{equation}
\begin{aligned}
G_m&=sigmoid(W_G^1v_{\phi}^P+W_G^2o_{m}^A),\\
h_{\mathscr{s},m}^A&=G_mo_{m}^A+(1-G_m)v_{\phi}^P,
\end{aligned}
\end{equation}
where $W_G^1$ and $W_G^2$ are two linear layer and $G_m$ is the gate weight. By adapting gated knowledge fusion, we can selectively integrate information from multi-source data and enhance the spatial correlation modeling and VFSRL in the active party.

\subsubsection{Cross-client Multi-level Alignment}\label{section:multi-level}
The cross-client VNA block employs multiple VNA modules for cross-client multi-level alignment. It receives the spatiotemporal representations $h_{ST}^P$ as input as set of multi-level representation from the passive party $P$ 
and conducts multi-level alignment with the inputs. This enables the passive parties to generate virtual nodes and the active party to conduct knowledge fusion with multiple VNA module for representation enhancement.
This design also allows a more comprehensive modeling of cross-client spatiotemporal dependencies at different granularity.
In addition, a level-to-level fusion is conducted within the multi-level alignment.
Specifically, the virtual node generated from representation $o_m^P$ at the passive party is fused with $o_{m+1}^A$ at the active party to derive $h_{\mathscr{s},m+1}^A$.
This incorporates the information from multiple clients at the lower layers of VFSRL at the active party, enabling a more effective spatial correlation modeling in the following spatial layers to enhance predictions.

\subsection{Collaborative Training}
\label{section:optimization}
The joint optimization of modules in HSTFL requires an exchange of embeddings and gradients between clients. In the forward process, the passive parties send virtual nodes to the active party for prediction. 
Then, the active party applies mean absolute error (MAE) as the loss function and generates gradients to update its local model privately in the backward process. Finally, the gradients of the virtual nodes will be sent back to the passive parties for them to update their local model privately. We provide a detailed complexity analysis of collaborative training with HSTFL in Appendix \ref{appendix:complexity}.

\section{Privacy Analysis}
\label{sec:privacy}
In this section, we discuss how to prevent attackers from reconstructing clients' private time series data in HSTFL. We follow the settings of~\cite{FIattack}, where the active party that holds more information aims to reconstruct the raw data of the passive party through attacks. First, we propose the evaluation of privacy leakage with attacks, and then demonstrate how we reduce the privacy leakage.

\subsection{Evaluation of Privacy Leakage}
\begin{myDef}[inverse function]
\textit{Given a function $f:D\to \mathbb{R}^d $ over a dataset $D$, the inverse function $f^{-1}$ is defined as}
\begin{equation}
    f^{-1}=argmin_{g}\sum_{x\in D}\|x-g(f(x))\|_2.
\end{equation}
\end{myDef}

The only information received by the active party are the virtual nodes from the passive parties. Therefore, We can regard the private model of a passive party as a function $f$, with input $x$ and output the virtual nodes $f(x)$. We formulate the attack function on HSTFL into $\mathscr{A}(v, \Theta)$ with the virtual nodes $v$ (\ie $f(x)$) and prior knowledge $\Theta$. An ideal attack algorithm would have $\mathscr{A}^{\Gamma}(\cdot, \Theta)=f^{-1}( \cdot )$, as the result of the inverse function will be close to the raw data. 

\begin{myDef}[information leakage]
\label{def:InfoLeak}
\textit{Given a function $f:D\to \mathbb{R}^d $ over a dataset $D$ and an attack function $\mathscr{A}$ with the prior knowledge $p$, the information leakage $\Lambda$ is defined as}
\begin{equation}
    \Lambda=\frac{1}{1+\frac{1}{|D|}\sum_{x\in D}{\|x-\mathscr{A}(f(x), \Theta)\|_2}}.
\end{equation}
\end{myDef}
 
As $\|x-\mathscr{A}(f(x), \Theta)\|_2$ reflects the closeness of the reconstructed input to the true input, Information Leakage (InfoLeak) $\Lambda$ is able to reflect privacy leakage of HSTFL with the attack function $\mathscr{A}$. InfoLeak equal to 1 means a perfect reconstruction, and being close to zero means a bad reconstruction. High InfoLeak means the framework is vulnerable to the attack, and vice versa. 

\subsection{Reducing Privacy Leakage in HSTFL}
We reduce privacy leakage in HSTFL on three aspects: reducing information leakage of virtual nodes through privacy-preserving framework construction, preventing perfect reconstruction with DP, and reducing prior knowledge leakage to defend against attacks.

\textbf{Privacy-preserving framework construction.}
HSTFL follows a no-model-sharing design of split learning ~\cite{vsplit}, where only data representations are exposed from the passive party during training and it is theoretically impossible to reconstruct the raw data without prior knowledge~\cite{splitnnSecure}. 
To resist attacks leveraging prior information, we strive to minimize unrelated information for predictions in virtual nodes by maximizing the local computation of clients without compromising the model's performance. This design can increase the difficulty of attacker to reconstruct raw data precisely as information of raw data are lost during forward~\cite{sfa}. 
Specifically, we adapt a temporal-then-spatial design for participants to process their time series data locally without exposing detailed information of time steps, and expose virtual nodes instead of raw embeddings of each time series to reduce information leakage $\Lambda$. 

\textbf{Differential privacy against perfect reconstruction.}
Differential privacy (DP) on the virtual nodes further ensures that attackers cannot perfectly reconstruct the raw data.

\begin{mythe}
\label{the:lower_bound}
    \textit{Given the lipschiz constant $L$ of the function $f$ at $x \in D$ with the noise generated by differential privacy $\mathscr{N}$ on virtual nodes, if $f(x)+\mathscr{N} \in f$, the distance between $x$ to the reconstructed data of the attack $\mathscr{A}^{\theta}(\cdot,\Theta)$ which achieves $\Lambda=1$ is bounded by $\frac{|\mathscr{N}|}{L}$.}
\end{mythe}

The proof of Theorem \ref{the:lower_bound} can be found in Appendix \ref{appendix:theorem}.
Theorem \ref{the:lower_bound} demonstrates that applying DP to the virtual nodes in HSTFL leads to deviations in the reconstructed data with the ideal attack. This prevents a perfect reconstruction that achieves $\Lambda=1$.

\textbf{Prior knowledge leakage reduction.}
Reducing the leakage of data-related information as prior knowledge $\Theta$ also contributes to defend against attacks. In HSTFL, the VNA module safeguards the geographic locations of the client's time series, and vertical federated spatiotemporal representation learning also protects the length of input time series and feature numbers for clients in HSTFL. The protection of these prior knowledge significantly increases the difficulty of carrying out attacks in real-world scenarios.

\begin{table*}[t]
  \setlength{\abovecaptionskip}{0.1cm}
  \centering
  \caption{ Overall results and ablation study of HSTFL with baselines on four real-world multi-source datasets, the best result of each task is presented in bold. }
  \label{tab:main_result}
  \begin{tabular}{c|ccc|ccc|ccc|ccc}
  \midrule
     \multirow{2}{*}{FL algorithm} & \multicolumn{3}{c|}{Lyon Parking}  & \multicolumn{3}{c|}{CHI Bike} &\multicolumn{3}{c|}{ Beijing Air Quality} &\multicolumn{3}{c}{NYC Bike} \\
     & MAE & RMSE& SMAPE  & MAE & RMSE& SMAPE & MAE & RMSE& SMAPE & MAE & RMSE& SMAPE\\
    \hline
     Local model & 11.522 & 17.314 & 0.0684 & 2.017 & 3.146 & 0.1853 & 19.178 & 25.627& 0.1226  & 2.007 & 2.940& 0.2300 \\
     Top1Sim    & 12.616 & 18.949 & 0.0717  & 2.344 & 3.657 & 0.2116 & 19.562 & 26.058 & 0.1191& 2.364 & 3.461 & 0.2661\\
     AVGSim     & 12.636 & 18.988 & 0.0721  & 2.366 & 3.690 & 0.2144 & 19.741 & 26.274 & 0.1199 & 2.300 & 3.377 & 0.2584\\
     FedSim      & 12.758 & 19.407 & 0.0714 & 2.248 & 3.568 & 0.1942 & 18.264& 24.410 & 0.1127 & 2.155 & 3.156 & 0.2415 \\
     FL-FDML     & 10.886 & 16.105 & 0.0673 &1.922 & 2.963 & 0.1796 & 19.018 & 24.876 & 0.1187 &2.007 &  2.936 & 0.2295 \\     
     FL-SplitNN      & 10.395 & 15.112 & 0.0649 & 1.884 & 2.892 &  0.1791 & 18.421 & 24.612 & 0.1134 & 1.984 &  2.898& 0.2271 \\
    \hline
     HSTFL-NoPVFSRL & 9.891 & 14.395 & 0.0632 & 1.821 & 2.763 &0.1760 & 18.273 & 24.360 &0.1138 & 1.886 & 2.725&0.2222 \\
     HSTFL-NoMLR& 10.151 & 14.912&0.0624  & 1.824 &  2.764 &0.1757  & 18.181 & 24.227 &0.1119 & 1.909 & 2.772& 0.2223 \\
    HSTFL-NoMA     & 9.722 & 14.013 & 0.0625 & 1.782 & 2.714 & 0.1734 & 17.583 & 23.472 & 0.1100 & 1.874 & 2.711 & 0.2205\\
    HSTFL-NoVNA & 9.944 & 14.094 & 0.0638 & 1.875 & 2.877 &0.1791 & 17.750 & 23.615 &0.1106& 1.964 & 2.872&0.2269 \\
    \hline
     HSTFL-DP & 10.216 & 14.892 & 0.0638 & 1.804 & 2.725 & 0.1756 & 17.849 & 23.891 & 0.1109 & 1.888 & 2.7430& 0.2212  \\
     HSTFL & \textbf{9.587} & \textbf{13.917} & \textbf{0.0614} & \textbf{1.757} & \textbf{ 2.647} & \textbf{0.1716} & \textbf{17.579} & \textbf{23.466}& \textbf{0.1094}  & \textbf{1.863} & \textbf{2.698}& \textbf{0.2189} \\
    \hline
  \end{tabular}
\end{table*}   

\section{Experiment}
\subsection{Experimental Setup}
\subsubsection{Datasets}
We use four multi-source spatiotemporal datasets to validate the effectiveness of HSTFL. 
\textbf{CHI Bike.} This dataset contains the demand data of bikes and taxis in Chicago, with taxi data enhancing bike demanded prediction as~\cite{costnet}. 
\textbf{Lyon Parking.} This dataset contains the parking availability data of parking lots and traffic flow in Lyon, with traffic flow data enhancing parking availability prediction as~\cite{citywideparking}.
\textbf{Beijing Air Quality.} This dataset contains air quality data and station-based weather data in Beijing, with weather data enhancing air quality prediction as~\cite{mastergnnc}.
\textbf{NYC Bike.} This dataset contains the demand data of bikes and taxis in New York, with taxi data enhancing bike demand prediction as~\cite{costnet}.
Details and preprocessing of these datasets are in Appendix \ref{appendix:preprocessing}. 


\subsubsection{Baselines} We compare HSTFL with six baselines to demonstrate the effectiveness of our algorithm.
\textbf{(1) Local model.} Local model is constructed by temporal representation learning and spatial representation learning. Local model does not utilize data from other clients in prediction.
\textbf{(2) FedSim~\cite{fedsim}.} FedSim matches cross-client data with high similarity with fuzzy linkage and conducts similarity based VFL. It processes input data individually, thus does not support intra-client data correlation modeling. We also selected its two variants, \textbf{Top1Sim} that match the item with the highest similarity and \textbf{AvgSim} that do not consider similarity in VFL for comparison. 
\textbf{(3) FL-FDML.} Feature Distributed Machine Learning (FDML)~\cite{fdml} is a vertical federated deep learning framework. We apply fuzzy linkage to match cross-client data in FL-FDML. 
\textbf{(4) FL-SplitNN.}  Split Neural Network (SplitNN)~\cite{vsplit} is another vertical federated deep learning framework. We apply fuzzy linkage to match cross-client data in FL-splitNN. 

We set $\delta=1e-4$ with $\epsilon=\infty $ by default for \textbf{HSTFL} and $\epsilon=8 $ for \textbf{HSTFL-DP}, which is the differential privacy version of HSTFL. To ensure a fair comparison, We adapt these algorithms to the spatiotemporal prediction tasks and strive to configure the same spatial and temporal modules within these VFL framework.

\subsubsection{Training Settings and Metrics.}
The learning rate for training HSTFL is $10^{-4}$ for Beijing Air Quality dataset and $10^{-3}$ for others. The weight decay is $10^{-4}$. The parameter $k$ for k-nearst neighbor is set to 5. We adapt Mean Absolute Error (MAE), Root Mean Squared Error (RMSE) and Symmetric Mean Absolute Percentage Error (SMAPE) to evaluate model performance, with InfoLeak (Definition \ref{def:InfoLeak}) and MAE to evaluate privacy leakage of HSTFL. More details about the training settings are placed in Appendix \ref{appendix:training}. 

\subsection{Overall Result}
We evaluate the performance of HSTFL on four real-world spatiotemporal prediction tasks. The result in Table \ref{tab:main_result} shows that HSTFL obtains superior performance on all four datasets. It outperforms the local model significantly, bringing about a 7.2\% to 16.8\% and a 8.21\% to 19.6\% improvement in MAE and in RMSE, respectively. It also shows an improvements in SMAPE across four datasets, which demonstrate that HSTFL is capable for integrating additional spatiotemporal data to enhance predictions. Besides, HSTFL shows superior performance to baselines. On one hand, FedSim, Top1Sim, and AVGSim lack the capability to model intra-client correlations in both two types of roles, resulting in a poor performance. On the other hand, FL-FDML and FL-SplitNN utilize fuzzy linkage for cross-client data matching. Though they bring improvements to the local models, they fail to comprehensively model the cross-client spatiotemporal correlations and perform worse than HSTFL. HSTFL excels these baselines by generating effective representations through VFSTRL and models cross-client spatiotemporal correlation with cross-client VNA block. In addition, HSTFL-DP still achieves good model performance, indicating that HSTFL is suitable for scenarios with strong privacy requirements.

\begin{table*}[t]
  \setlength{\abovecaptionskip}{0.1cm}
  \centering
  \caption{ Privacy evaluation on four real-world multi-source datasets with inference attacks.  $\uparrow$ means bigger value is better while $\downarrow$ means smaller value is better. The best results of each attack are presented in bold.}
  \label{tab:privacy_result}
  \begin{tabular}{c|c|cc|cc|cc|cc}
  \midrule
     \multirow{2}{*}{FL algorithm} &  \multirow{2}{*}{Attack method} & \multicolumn{2}{c|}{Lyon Parking}  & \multicolumn{2}{c|}{CHI Bike} &\multicolumn{2}{c|}{ Beijing Air Quality} &\multicolumn{2}{c}{NYC Bike} \\
    
    &  & InfoLeak $\downarrow$ & MAE $\uparrow$ & InfoLeak $\downarrow$& MAE $\uparrow$ & InfoLeak $\downarrow$& MAE $\uparrow$ & InfoLeak $\downarrow$&  MAE $\uparrow$ \\
    \hline
      -&\emph{Mean}&         0.4885 &0.7871  & 0.5287& 0.3696  & 0.5252 & 0.7546& 0.5211 & 0.7393 \\
      -&\emph{Random Guess}&  0.4767 & 0.8971 & 0.4765 &0.7937 &  0.4836 &0.8901 &  0.4688 & 0.9907  \\
    \hline
    HSTFL-NoVNA &  White-box attack & 0.7272 &  0.1722 &0.5875 & 0.1508 &  0.6146 &0.4386 & 0.6594 & 0.2830 \\
    HSTFL& White-box attack & \textbf{0.5780} &\textbf{0.4951} &   \textbf{0.5453} & \textbf{0.3209} & \textbf{0.5598} &\textbf{0.6035} &  \textbf{0.5796} & \textbf{0.6756} \\
    \hline
    HSTFL-NoVNA&  Query-free attack & 0.5784& 0.4829  & 0.5103 &  0.4845 &0.4939 &0.8172  & 0.5102 &  0.7152\\
    HSTFL & Query-free attack&  \textbf{0.3724} & \textbf{1.2812} & \textbf{0.4538} &\textbf{0.7241} &  \textbf{0.3386} &\textbf{1.5228}  & \textbf{0.5079}& \textbf{0.7262} \\
    \hline
  \end{tabular}
  \vspace{-0.3cm}
\end{table*}

\subsection{Ablation Study}
\subsubsection{Effect of VFSTRL}
To verify the effectiveness of VFSTRL, we proposed two variants of HSTFL including \textbf{(1) HSTFL-NoPVFSRL}: VFSRL is conducted only in the active party but not the passive parties. \textbf{(2) HSTFL-NoMLR}: the spatiotemporal representation is not a multi-level representation. The output of the topmost layer of the passive party's private model is selected as spatiotemporal representation. Experiment results in Table \ref{tab:main_result} show that HSTFL consistently outperforms these two variants. This result demonstrates the significance of modeling intra-client spatiotemporal correlations and generating multi-level representations for better cross-client spatiotemporal correlation modeling and predictions.

\subsubsection{Effect of the Cross-client VNA Block}
We proposed two variants of HSTFL to demonstrate the effectiveness of the cross-client VNA Block. \textbf{(1) HSTFL-NoVNA}: the VNA modules are replaced with fuzzy linkage to match cross-client time series with locations, \textbf{(2) HSTFL-NoMA}: only one VNA module is applied for multi-level alignment. Results in Table \ref{tab:main_result} show that HSTFL consistently performs better, showing that the VNA module effectively enhance the modeling of the cross-client data correlations and the multi-level alignment provides more information with different granularity.

\begin{figure}
  \centering
  \includegraphics[width=0.4\textwidth]{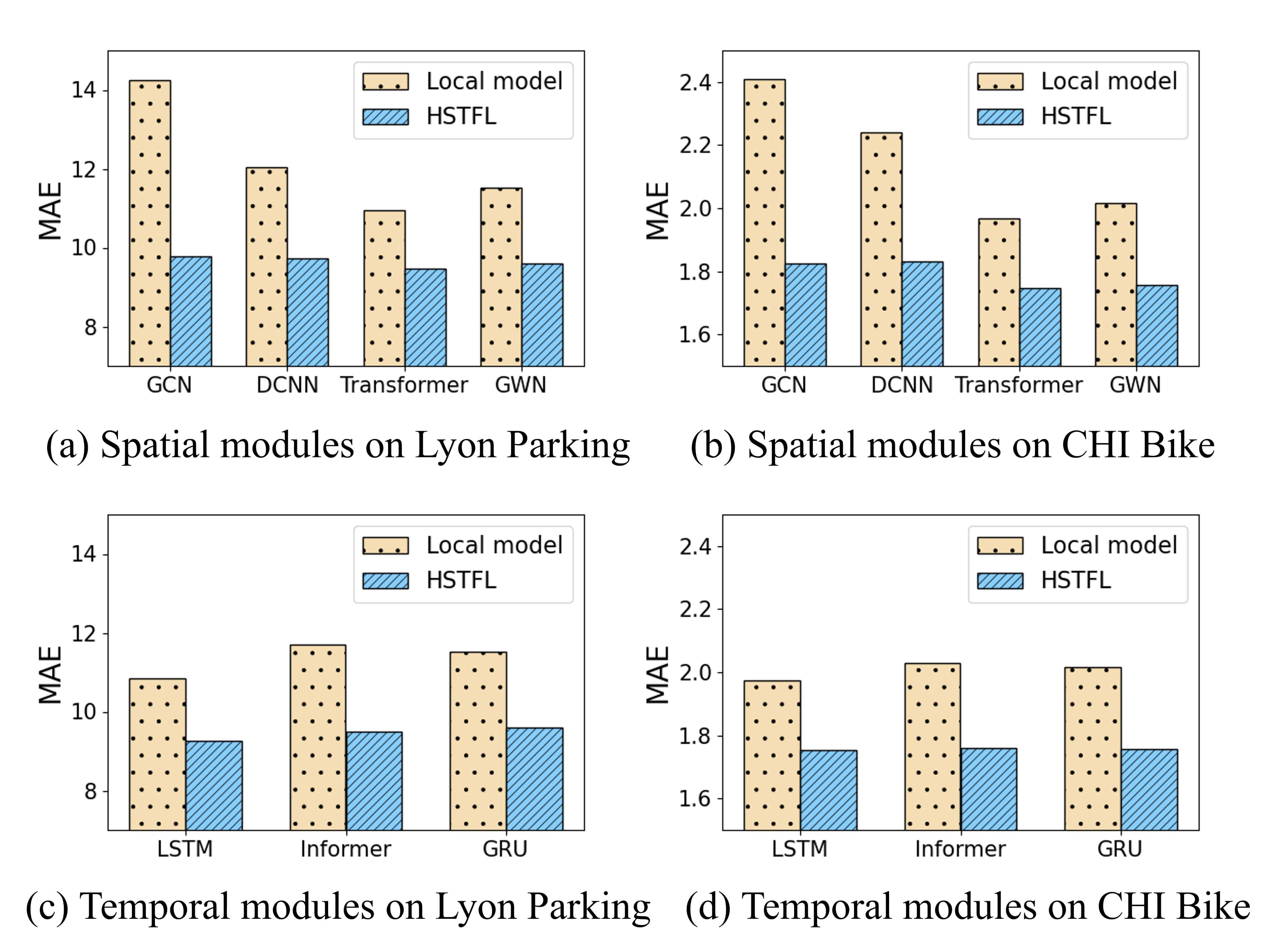}
  \caption{HSTFL with different local spatiotemporal module.}
  \label{fig:STmodule}
\end{figure}

\subsection{Module Agnostic Evaluation}
In this subsection, we evaluate the performance of HSTFL and local models while modifying the local spatiotemporal modules within the framework. We compare the effectiveness of spatial layers (GWN~\cite{gwnet} by default) with three different spatial modules: Graph Convolution (GCN)~\cite{gcn}, Diffusion Convolution (DCNN)~\cite{dcrnn}, Transformer~\cite{lightcts}. We also compare temporal layers (GRU by default) with two different modules: LSTM and Informer~\cite{informer}.

From Figure \ref{fig:STmodule} we can see that HSTFL can consistently outperform local models with whatever spatial and temporal modules. Despite the superior performance of certain modules, HSTFL successfully leverages its power to enhance performance. This demonstrates that HSTFL provides clients with the flexibility to arbitrarily select the spatiotemporal module according to their local data. 

\subsection{Privacy Evaluation}
In this subsection, we demonstrate that HSTFL is able to defend against inference attacks with prior knowledge.
\subsubsection{Evaluation Methods}
We employed two inference attacks $\mathscr{A}$ on data representations to evaluate the privacy leakage of HSTFL: White-box attack and Query-free (black-box) attack~\cite{MIattackC,MIattackJ}.
The prior knowledge for these algorithms are the passive parties' private model for the White-box attack and a similar dataset for training the private model for the Query-free attack. 
In addition, we apply \emph{Mean} and \emph{Random guess}, which are two attacks that do not rely on data representations as baseline (The result of \emph{Mean} and \emph{Random guess} only depends on the dataset). \emph{Mean} assume all values in the time series are the mean value. \emph{Random Guess} that applies a uniform distribution to guess the values in the time series. The implementation of these attacks are reported in the Appendix \ref{appendix:attackalg}.

\subsubsection{Performance Against Attack}
Table~\ref{tab:privacy_result} shows the overall performance of HSTFL against attack. InfoLeak represents the information leakage of the framework with the attack, while MAE represents the distance between the scaled reconstructed data and the scaled real data. The result of Query-free attack is worse than \emph{Mean} and \emph{Random Guess}, which suggests that HSTFL shows resilience against attacks as the attackers are insufficient to conduct effective attacks even with some prior knowledge. Although the White-box attacks achieve better than \emph{Mean} and \emph{Random Guess}, the effectiveness of the attacks are not significantly superior to them. The overall privacy leakage of HSTFL is limited.

We also demonstrate the effectiveness of VNA in reducing privacy leakages, and HSTFL shows better results against inference attacks with HSTFL-NoVNA. This indicates that the exposure of representations for time series in fuzzy linkage VFL ~\cite{fedsim} increases the risk of privacy leakage, and further demonstrates the effectiveness of VNA's privacy protection in HSTFL.

\begin{figure}
  \centering
  \includegraphics[width=0.36\textwidth]{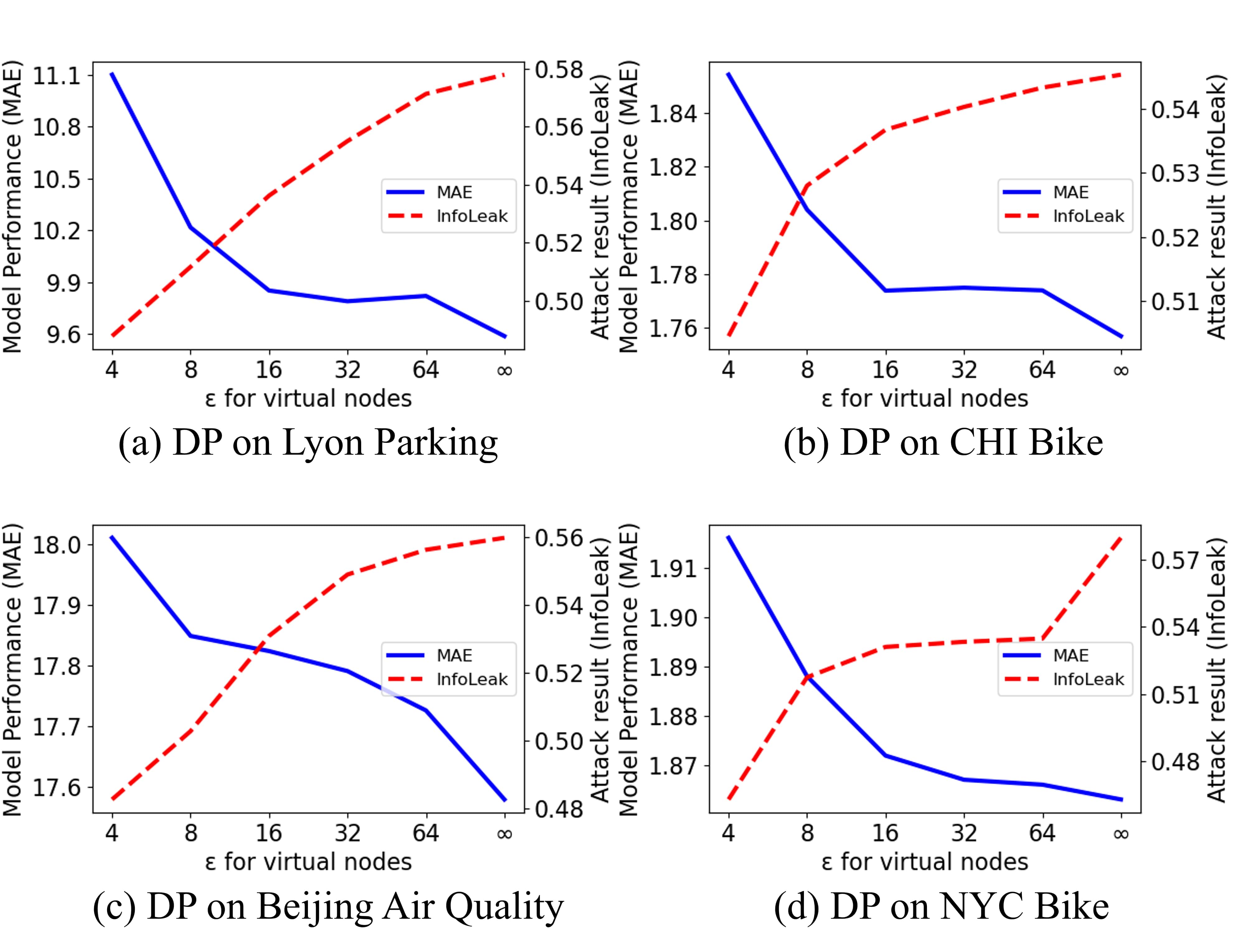}
  \caption{The model performance and attack result of the White-box attack with differential privacy.}
  \label{fig:dp-tradeoff}
\end{figure}

\subsubsection{Trade-off with Differential Privacy}
\label{section:dp}
Additionally, we vary $\epsilon=\{4, 8, 16, 32, 64, \infty\} $ to show how differential privacy affects the model performance and data privacy in HSTFL. Figure \ref{fig:dp-tradeoff} demonstrates the result of white-box attack on HSTFL with different levels of differential privacy. The model's performance decreases as $\epsilon$ decreases, but the attack results also worsen. Thus, there is a trade-off between accuracy and privacy, where a smaller $\epsilon$ value adds more noise to virtual nodes for stronger privacy protection but leads to more prediction error. Noteworthy, a small value of $\epsilon$ do not completely ruin the model's performance as the differential privacy is only applied on the virtual nodes, while the information of the active party remains complete.

\section{Conclusion}
This paper proposed Heterogeneous SpatioTemporal Federated Learning (HSTFL), a privacy-preserving machine learning framework that enables multi-party collaborative spatiotemporal forecasting without direct access to multi-source private data. 
To handle cross-domain feature heterogeneity and cross-client geographical heterogeneity with privacy-preserving, we proposed vertical federated spatiotemporal representation learning and cross-client virtual node alignment block to model the spatiotemporal correlations of heterogeneous data. Experiment shows that HSTFL effectively resists inference attacks with prior knowledge and significantly improves the performance of spatiotemporal predictions by incorporating multi-source private spatiotemporal data.

\bibliographystyle{ACM-Reference-Format}
\bibliography{sample-base}

\clearpage
\appendix
\section{Complexity of HSTFL}
\label{appendix:complexity}
The cost and complexity of HSTFL mainly come from two parts: computational and communication.

The computational complexity in HSTFL stems from the representation learning in VFSTRL and cross-client spatiotemporal modeling in cross-client VNA blocks, which are manageable. In VFSTRL, the computational burden is related to the selected modules, length of the time series, and the number of layers in spatiotemporal module. In addition, the computational cost in the VNA section is controlled by the number of layers in the Spatial layer. HSTFL employs a flexible design in HSTFL, which allows clients to freely choose local modules and parameters to balance computational burden and prediction effectiveness.

The communication complexity of HSTFL is related to the number of forwards and backwards and mainly arises from the cross-VNA blocks, where virtual nodes and gradient are exchanged between the passive parties and the active party. In the forward process, the communication cost brought by the prediction of each item is $L \times N \times H$ ( $L$ is the number of VNA modules in the cross-client VNA block, $N$ is the number of time series in the active party, and $H$ is the dimension of the virtual node). By summing then transforming the immediate representations of the three aggregation instead of first concatenating them, HSTFL can reduce communication cost without sacrificing performance. In the backward process, these virtual nodes will receive a gradient the gradients of the same size. Hence, the communication complexity of a forward or backward in HSTFL is $\mathcal{O}(LNH)$.

As a cross-silo FL algorithm, clients in HSTFL are typically institutions or companies, enabling them to afford such computational effort and communication cost. Hence, the complexity enables a good deployment of HSTFL in real-world scenarios.

\section{Proof of Theorem \ref{the:lower_bound}}
In this section, we first analyze the characteristic of the functions with ideal attack, then demonstrate why differential privacy is able to prevent perfect reconstruction. 

\label{appendix:theorem}
\begin{mylem}
\label{the:bijection}
    \textit{If InfoLeak equals to $1$ with the attack function $\mathscr{A}^{\theta}$, the function $f(\cdot)$ is bijection.}
\end{mylem}
\textit{Proof. }  By contradiction. If the function $f(\cdot)$ is not bijection, there are $x,y\in D$ and $x \neq y$, but $f(x)=f(y)$ and $\mathscr{A}^{\theta}(f(x))=\mathscr{A}^{\theta}(f(y))$. This is a contradiction as the perfect reconstruction requires both $x=\mathscr{A}^{\theta}(f(x))$ and $y=\mathscr{A}^{\theta}(f(y))$ to achieve $\Lambda=1$. Therefore, the function $f(\cdot)$ must be is bijection. 


\setcounter{mythe}{1}    
\begin{mythe}
\textit{Given the lipschiz constant $L$ of the function $f$ at $x \in D$ with the noise generated by differential privacy $\mathscr{N}$ on virtual nodes, if $f(x)+\mathscr{N} \in f$, the distance between $x$ to the reconstructed data of the attack $\mathscr{A}^{\theta}(\cdot,\Theta)$ which achieves $\Lambda=1$ is bounded by $\frac{|\mathscr{N}|}{L}$.}
\end{mythe}

\textit{Proof. }  By Lemma \ref{the:bijection}, we have for $x\in D$ and $\mathscr{v} \in f$, $\mathscr{A}^{\theta}(f(x),\Theta)=x$ and $f(\mathscr{A}^{\theta}(\mathscr{v},\Theta))=\mathscr{v}$. From the Lipschitz continuous, 
\begin{equation}
    |x-\mathscr{A}^{\theta}(f(x)+\mathscr{N}, \Theta)|\geq \frac{|f(x)-(f(x)+\mathscr{N})|}{L}=\frac{|\mathscr{N}|}{L}.
\end{equation}

Hence, there must be a deviation between the raw data and reconstructed data of the ideal attack.

\section{Details of Experiment}
\subsection{Datasets}
\label{appendix:preprocessing}

We used four multi-source real-world datasets in the experiment section, namely \textbf{CHI Bike}, \textbf{Lyon Parking}, \textbf{Beijing Air Quality}, and \textbf{NYC Bike}. The processed data is summarized in the Table \ref{tab:dataset}.

\subsubsection{Details of datasets}
We collect the data from different source and select different temporal span for constructing datasets. 

For CHI Bike dataset, the bike data of Chicago is from Divvybikes\footnote{\url{https://divvybikes.com/system-data}} and the taxi data is from Cityofchicago \footnote{\url{https://data.cityofchicago.org/Transportation/Taxi-Trips/wrvz-psew/about_data}}. We selected data from 04/01/2016 to 06/30/2016 to construct the dataset. The prediction task of this dataset is to forecast the pick-up and drop-off demand of bike in each region as ~\cite{costnet,mdtp}. 

For Lyon Parking dataset, the parking availability data and traffic data in Lyon (France) are collected from Data.Grandlyon\footnote{\url{https://data.grandlyon.com/portail/fr/accueil}}. The website provided us with data from 04/01/2023 to 06/30/2023 to construct datasets for experiments. The prediction task of this dataset is to forecast the parking availability of each parking lots as ~\cite{citywideparking}. 

For Beijing Air Quality dataset, both the air quality and weather are from KDD CUP 2018 \footnote{\url{https://www.biendata.xyz/competition/kdd}}, it includes the weather and air quality data of Beijing from 01/01/2017 to 12/31/2017. The prediction task of this dataset is to forecast the Air Quality Index (AQI) for air quality prediction at each air quality station, which is derived by the Chinese AQI standard as ~\cite{mastergnnc}. 

For NYC Bike dataset, the bike and taxi data are collected from Citybikenyc\footnote{\url{https://citibikenyc.com/system-data}} and Nyc.gov\footnote{\url{https://www.nyc.gov/site/tlc/about/tlc-trip-record-data.page}} respectively. We selected data from 04/01/2016 to 06/30/2016 to construct the dataset. The prediction task of this dataset is to forecast the pick-up and drop-off demand of bike in each bike station as ~\cite{costnet,mdtp}. 

\subsubsection{Preprocessing}
We counted the pick-up record and drop-off record of each bike station and taxi zone during different time periods to construct the multi-source spatiotemporal dataset for CHI Bike and NYC Bike datasets. The default sampling rate of these two dataset is 30 minutes (each time slot represents 30 minutes).
We average the parking availability of parking lots in the Lyon Parking dataset in each time slot, and sum the traffic flow in the time slots. The default sampling rate of Lyon Parking is 15 minutes.
For Beijing Air Quality dataset, we just record the time series normally. The default sampling rate of Beijing Air Quality is 1 hour.

We use linear interpolation for missing values in the data. We also preprocess these datasets by removing time series with a high number of missing values or those whose locations are too far from the cluster center. Additionally, we applied grid-based processing to the taxi and bike data of CHI Bike to increase the data volume as ~\cite{costnet,mdtp}. 

All datasets followed an 8:1:1 ratio for the train, validation, and test. It is noteworthy that, we divided the entire year of the Beijing Air Quality dataset into four segments of three months representing four seasons. Each segment was further divided into train, validation, and test sets to minimize distribution differences between them as~\cite{mastergnnc}.

\begin{table*}[htbp]
    \centering
    \caption{Dataset description. $|N|$ and $f_{in}$ represents the number of time series and its feature respectively. $f_{out}$ represents the feature of the time series for predictions. $A$ and $P$ represents the active party and passive party respectively.}
    \label{tab:dataset}
     \renewcommand{\arraystretch}{1.5}
    \begin{tabular}{|c|c|c|c|c|c|c|c|c|c|}
        \hline
        Dataset name & Data of $A$ &$|N^A|$ & $f_{in}^A$ & $f_{out}$ & Data of $P$ & $|N^P|$ & $f_{in}^P$ & Temporal span & Sampling rate\\
        \hline
        CHI Bike  & Bike demand& 53 & 2 & 2 & Taxi demand& 77 & 2 & 3 month &30 minutes \\
        \hline
        Lyon Parking  & Parking availability & 36 & 2 & 1 & Traffic flow& 42 & 2 & 3 month &15 minutes\\
        \hline
        Beijing Air Quality  &Air quality& 35 & 7 & 1 & Weather& 651 & 5 & 1 year&60 minutes \\
        \hline
        NYC Bike  & Bike demand& 285 & 2 & 2 & Taxi demand& 76 & 2 & 3 month &30 minutes\\
        \hline
    \end{tabular}
\end{table*}

\subsection{Training Details of HSTFL}
\label{appendix:training}
The history time steps of the active party is 12 for CHI Bike, NYC Bike and Lyon Parking and 48 for Beijing Air Quality. The future time steps (\ie $\tau$) of the active party in all dataset are set to 12. We adjust the historical time steps for the passive party to achieve the same temporal span in the active party by default (if the time intervals are the same in the active party and the passive party, the historical time steps are the same).
Each numerical feature of dataset in HSTFL is normalized by a standard scaler. The hidden size of embedding layer, temporal module and spatial module are selected from $\{32,64,128\}$. The optimizer is Adam optimizer. All the models are trained with MAE loss, with max epoch 250.

In addition, to ensure compatibility of the spatiotemporal model across clients with varying node counts, we employed DCNN as the spatial kernel in~\ref{sec:fedavg}. Furthermore, we reorganized the dataset based on an extended history of time steps in ~\ref{sec:timeseries} for a fair comparison.

\begin{figure}
  \setlength{\abovecaptionskip}{0.1cm}
  \centering
  \includegraphics[width=0.47\textwidth]{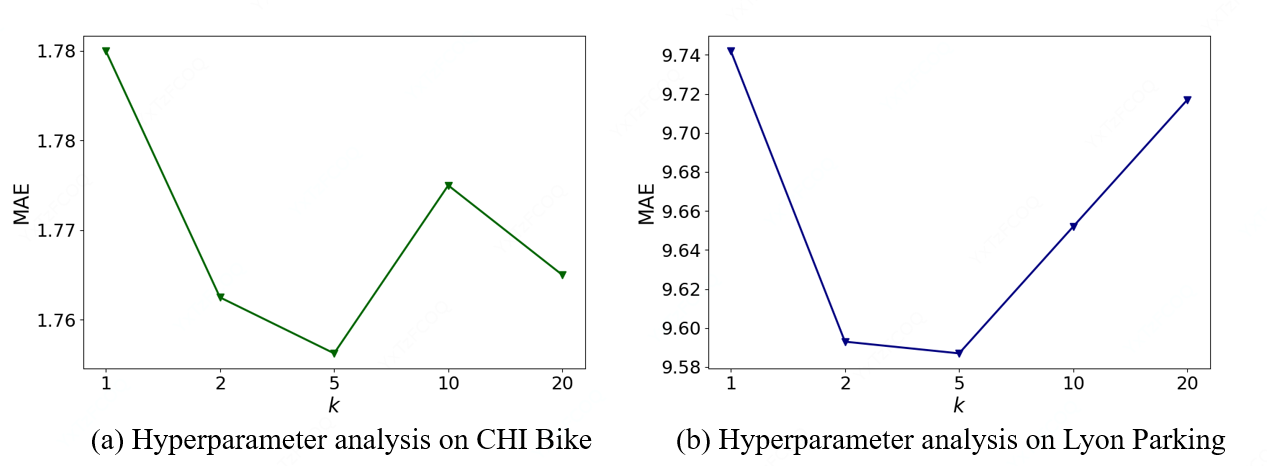}
  \caption{Performance of HSTFL with different $k$}
  \label{fig:hyperparam}

\end{figure}

\section{Additional Experimental Results}
\subsection{Hyperparameter Analysis}
We provide an analysis of the hyperparameter here in HSTFL. In addition to the hyperparatmeter of differential privacy that has been demonstrate in section \ref{section:dp}, the only parameter of HSTFL is $k$ for KNN in cross-client VNA block. This hyperparameter determines how HSTFL will modeling cross-client spatiotemporal correlations with the awareness of spatial proximity.

We demonstrate the effect of varying $k$ in Figure ~\ref{fig:hyperparam}. The result on CHI Bike and Lyon Parking dataset shows that an excessively low values of $k$ lead to a slight decrease in performance. This indicates that focusing on an extremely narrow range of spatial proximity information is inappropriate.
Although a larger $k$ is advantageous for capturing a wider range of spatial proximity information, it may dilute essential local information. As the adaptive aggregation in VNA is designed to modeling long-distance and implicit correlation, setting a extremely high $k$ is also inappropriate.
When $k$ is around 5, HSTFL performs best on the CHI Bike and Lyon Parking datasets.

\subsection{Effect of Heterogeneous Features}
\label{sec:fedavg}
Clients in HSTFL holds completely different types of data with heterogeneous data features. This significantly undermine the effectiveness of the HFL algorithm that relys on homogeneity of data across clients.
On one hand, heterogeneous data collected by different parties may have varying dimensions, which prevents the model from being trained normally across different clients. On the other hand, the data distribution of heterogeneous data features differs significantly across clients, which affect the learning of the federated model.

\begin{figure}
  \setlength{\abovecaptionskip}{0.1cm}
  \centering
  \includegraphics[width=0.47\textwidth]{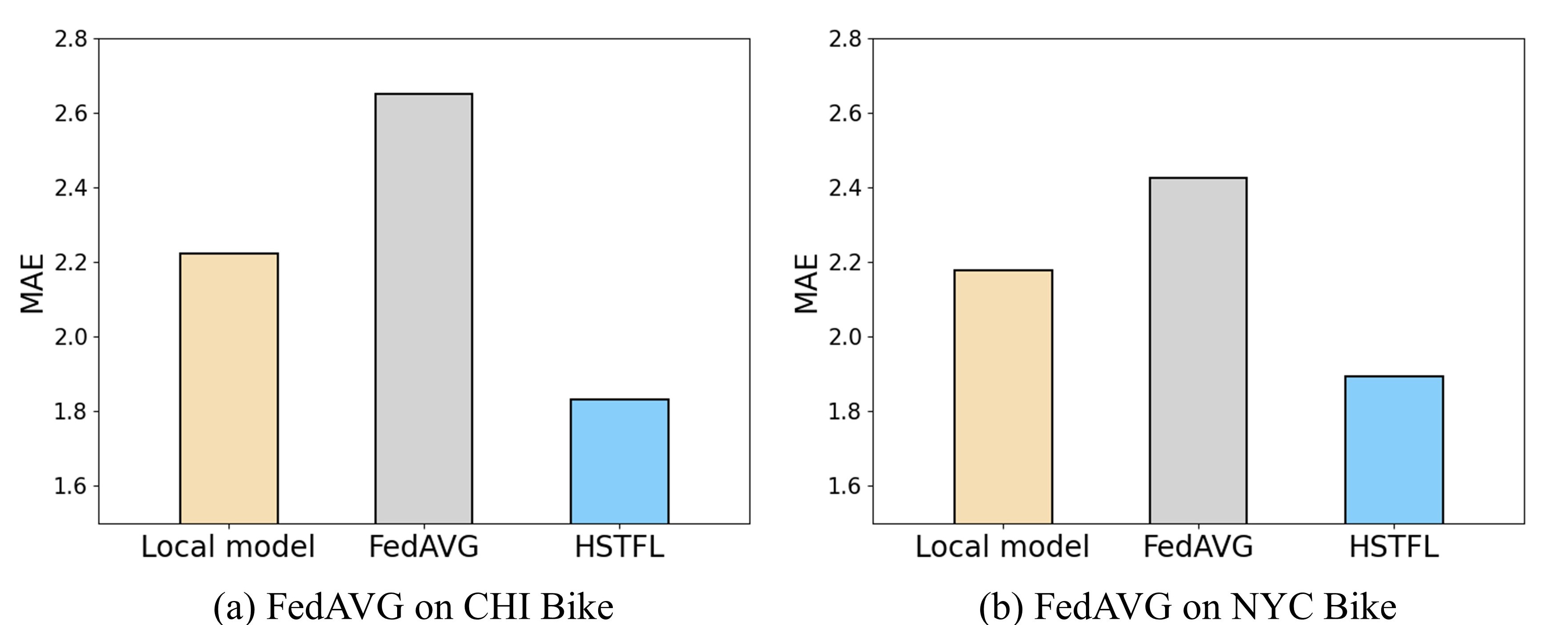}
  \caption{Effect of heterogeneous features}
  \label{fig:hfl}

\end{figure}

We conduct experiment of local model, FedAVG and HSTFL with CHI Bike and NYC Bike dataset. The clients hold traffic demand data for taxis and bikes that share similarities in terms of type and data distribution ~\cite{costnet,mdtp} with these two datasets.
As shown in Figure ~\ref{fig:hfl}, though the clients hold a similar type of data, the heterogeneity between bike demand and taxi demand undermines the performance of the HFL algorithm. The incorporation of additional data with FedAVG even leads to worse performance compared to the local model. In contrast, HSTFL effectively handles both cross-domain feature heterogeneity and cross-client geographical heterogeneity in the data, achieving a good performance.

\begin{figure}
  \setlength{\abovecaptionskip}{0.1cm}
  \centering
  \includegraphics[width=0.47\textwidth]{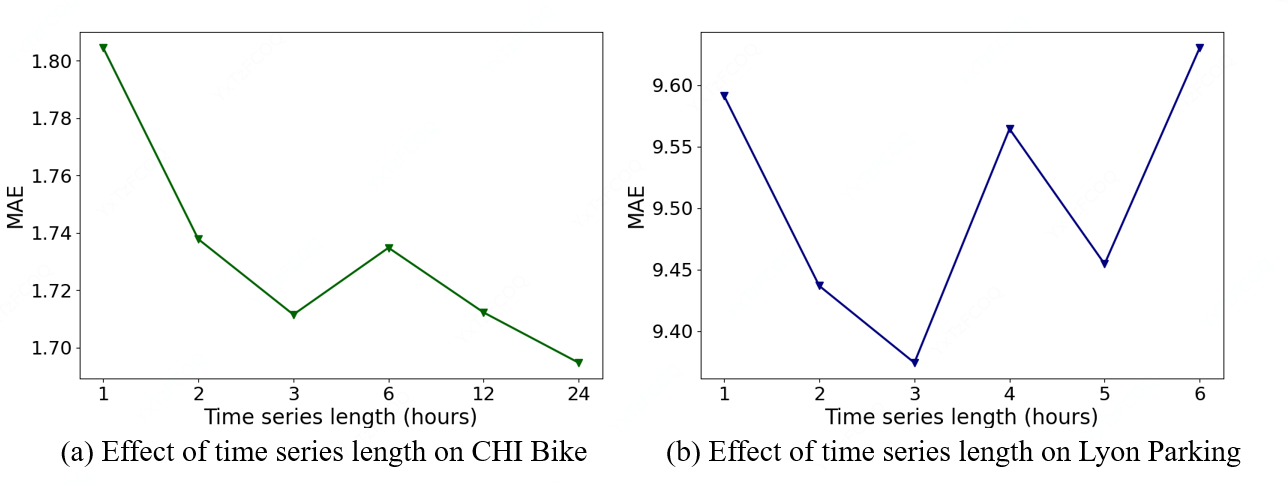}
  \caption{Effect of time series length}
  \label{fig:tl}

\end{figure}
\begin{figure}
  \setlength{\abovecaptionskip}{0.1cm}
  \centering
  \includegraphics[width=0.47\textwidth]{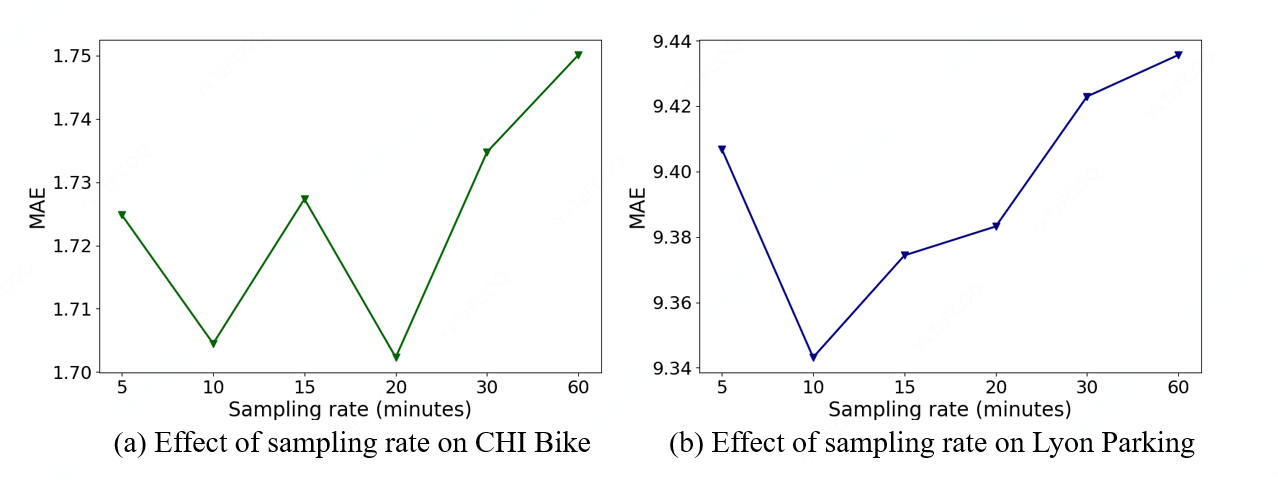}
  \caption{Effect of sampling rate}
  \label{fig:ts}

\end{figure}

\subsection{Effect of Heterogeneous Time Series}
\label{sec:timeseries}
In this section, we demonstrate the performance of HSTFL on time series with different length and sampling rate among clients. 
We re-segmented the CHI Bike and Lyon Parking datasets to show how HSTFL performs when faced with time series of different lengths or sampling rates from passive parties, while keeping the active party's time series length and sampling rate constant.

As shown in Figure ~\ref{fig:tl} and Figure ~\ref{fig:ts}, both the length and sampling rate of time series have an strong impact on the prediction on HSTFL.
A longer time series length and a shorter sampling rate may contain more information for better spatiotemporal prediction, but the gains from extended time series are also influenced by the dataset, prediction task, and the spatiotemporal module that employed. 
Additionally, excessively short sampling rates and time series lengths can lead to an increased computational burden.
Considering these issues, we adopted a time-then-graph architecture with a flexible temporal representation learning in HSTFL. This enable clients in HSTFL to use time series with different lengths or sampling rates for enhanced prediction with heterogeneous modules.
This flexibility also lowers the barrier for privacy-preserving multi-party collaborative spatiotemporal forecasting, enabling more potential participants to join HSTFL.

\section{Details of Privacy Evaluation}
\label{appendix:attackalg}
\subsection{Attack Algorithm for Evaluation}
Urban spatiotemporal models can be vulnerable to attacks~\cite{fan}. To address this concern, we employed two inference attack on data representations, the White-box attack and Query-free (black-box) attack\cite{MIattackC,MIattackJ}, aiming to demonstrate its ability to provide effective privacy protection. These two attacks are server as the attack method $\mathscr{A}$ to evaluate the privacy leakage of HSTFL and they require certain hard-to-obtain prior knowledge in the real-world to assist in the attacks. We have made adjustments on these attacks to apply them on spatiotemporal data. 

\subsubsection{White-box Attack}
The White-box attack is conducted during the inference stage of HSTFL and it requires the passive party's well-trained local model with parameters (local model needs to be obtained through model stealing). With this prior knowledge, the White-box attack generates representations (\ie virtual nodes in HSTFL) by inputting reconstructed data and optimizes this reconstructed data with gradient descent to make the generated embeddings as close as possible to the real embeddings (\ie virtual nodes in HSTFL). Given the passive party's model parameter $f_\theta$ with the data representation $z$, the White-box attack is defined as
\begin{equation}
    ED(x,z,f_\theta)=||z-f_\theta(x)||_2^2,    
\end{equation}
\begin{equation}
        TV(x)=\sum_{i,j}(|x_{i+1,j}-x_{i,j}|^2+|x_{i,j+1}-x_{i,j}|^2)^{\beta/2},
\end{equation}
\begin{equation}
    x^{*}=argmin_x \quad ED(x,z,f_\theta)+\lambda TV(x).
\end{equation}
The TV loss are designed to smooth the generated image data by reducing drastic fluctuations to improve data reconstruction. $\beta$ here is a hyperparameter to control the importance of TV loss. To adapt the attack on the time series, we modify the two-dimensional TV loss for image data into one-dimension TV loss for smoothing the time-series as
\begin{equation}
    TV_{T}(x)=\sum_{i}(|x_{i+1}-x_{i}|^2)^{\beta/2}.
\end{equation}
Then the objective of White-box attack on HSTFL is 
\begin{equation}
    x^{*}=argmin_x \quad ED(x,z,f_\theta)+\lambda  TV_{T}(x).
\end{equation}

Ideally, the reconstructed data would closely resemble the real data, but the embedding itself may not contain all the information of the raw data~\cite{imagerepresentation}. Therefore, the effectiveness of the attack is also constrained by the method of generating the embeddings.

\subsubsection{Query-free Attack}
The Query-free attack is also conducted during the inference stage of model and it requires a batch of data that is close to the training data of the passive party model, which can be obtained through data collection. The specific procedures of the Query-Free attack involves generating a new model $f_{\hat{\theta}}$ that is close to the passive party model $f_\theta$ with this batch of training data $\{\hat{x},\hat{y}\}$, and then conducting the White-box attack on this new model to recover the original data. Given the model of the active party's $f_{A}$, the objective of the Query-free attack is,
\begin{equation}
    f_{\hat{\theta}}=argmin_g Loss(f_{A}(g(\hat{x})),\hat{y}),
\end{equation}
\begin{equation}
    x^{*}=argmin_x \quad ED(x,z,f_{\hat{\theta}})+\lambda TV_{T}(x).
\end{equation}
Ideally, the new model will be close to the passive party's private model, but it depends on the distribution differences between the real training data and the new batch of data.

As the prior knowledge of the White-box attack is consider stronger than the Query-free attack, we can consider the White-box attack as a stronger attack algorithm, while Query-free attack represents a weaker attack algorithm. In addition, we also proposed two guessing-based attack method as baseline, \emph{Mean} and \emph{Random guess}. These two attack methods do not need to access the embeddings (\ie virtual nodes in HSTFL), and therefore, their results are only correlation to the characteristics of the dataset itself and independent of the training methods and frameworks. The \emph{Mean} refers to the attacker guessing that all values in the time series are equal to the mean of that feature, while the \emph{Random guess} refers to the attacker using a Gaussian distribution and random numbers to guess all values in the time series. These two methods are used to assess the effectiveness of an attack. If the performance of an attack is weaker than these two methods, it can be considered ineffective because it cannot extract meaningful information from the embeddings (\ie virtual nodes in HSTFL).

To carry out the attacks effectively, we assume that all four attack methods holds the standard scalers for data normalization. For the two inference attacks on spatiotemporal data, this information is obtained through stealing and statistical analysis of the new batch of data. In the case of the two guessing-based attacks, the standard scaler is given as prior knowledge for the attacks.

\subsection{Attack Implementations}
The loss function of the inference attacks is set to mean square error (MSE), $\lambda$ and  $\beta$ for TV loss is set to  $10^{-4}$ and $2$, respectively. Considering that both the White-box attacks and the Query-free attacks are methods with huge computational burden that aim to reconstruct individual samples, we employed K-means to select a representative batch of time series from the test set as the attack target. The learning rate is set to $0.1$, the weight decay is set to $10^{-5}$ and the training epoch of attack is set for 500. 

In the White-box attack, the private model of the passive parties is given to the attacker to start the attack. In the query-free attack, the training set and the validation set are given to the attack to reconstruct the passive party's model. The attacker will optimize the model with the training set and conduct validation on the validation set. The attack experiments of a dataset are enact on different models with same batch of data in the repeated experiments.

\section{Limitations}
\label{appendix:limitations}
HSTFL is a cross-silo spatiotemporal federated learning algorithm that enables collaborative spatiotemporal forecasting among multiple parties without direct access to their private data. Its limitations include: (1) HSTFL focuses on modeling the spatiotemporal correlations among heterogeneous data, thus it is not expected to perform well in non-spatiotemporal tasks. (2) As a cross-silo spatiotemporal FL algorithm, HSTFL cannot address privacy concerns arising from data collection processes in clients like companies and institutions. (3) The performance improvement of HSTFL relies on the correlations between multi-source spatiotemporal data. It may not be effective for all types of multi-source spatiotemporal data, nor can it be used to select effective multi-source spatiotemporal data for prediction. We plan to address these limitations to improve the generality and effectiveness of the HSTFL framework in the future.

\end{document}